\pdfoutput=1

\documentclass[11pt]{article}

\usepackage{ACL2023}

\usepackage{times}
\usepackage{latexsym}

\usepackage[T1]{fontenc}

\usepackage[utf8]{inputenc}

\usepackage{microtype}

\usepackage{inconsolata}
\usepackage{amsmath,amsfonts,bm}

\usepackage{times}
\usepackage{latexsym}

\usepackage{url}
\usepackage{booktabs}
\usepackage{multirow}
\usepackage{array, caption, floatrow, makecell, booktabs}
\usepackage{wrapfig}
\usepackage{floatrow}
\usepackage{comment}
\usepackage{enumitem}
\usepackage{accents}
\usepackage{xspace,mfirstuc,tabulary}
\usepackage{graphicx}
\usepackage{color}
\usepackage{subcaption}

\DeclareMathOperator*{\argmax}{argmax}

\newcommand{\uvbert}{U-VisualBERT\xspace}
\newcommand{\muvla}{$\mu$-VLA\xspace}
\newcommand{\vlmixer}{VLMixer\xspace}
\newcommand{\eteuvlp}{E2E-UVLP\xspace}
\newcommand{\modelname}{RELIT\xspace}
\newcommand{\modelnamefull}{\textbf{R}elative r\textbf{E}presentation-based \textbf{L}anguage-\textbf{I}mage pre-\textbf{T}raining\xspace}
\newcommand{\tagretrv}{Retrv~(Tag)\xspace}
\newcommand{\relrepretrv}{Retrv~(Relrep)\xspace}
\newcommand{\nlvr}{$\text{NLVR}^2$\xspace}

\title{Weakly Supervised Vision-and-Language Pre-training with\\ Relative Representations}

\author{
 Chi Chen$^{1,3}$,
 Peng Li$^{*\:2,4}$,
 Maosong Sun\thanks{\:\:Corresponding authors: Peng Li (lipeng@air.tsinghua.\\edu.cn) and Maosong Sun (sms@tsinghua.edu.cn)}$^{\:\:\:1,3}$, Yang Liu$^{1,2,3,4}$\\
 $^1$Dept. of Comp. Sci. \& Tech., Institute for AI, Tsinghua University, Beijing, China\\
 $^2$Institute for AI Industry Research (AIR), Tsinghua University, Beijing, China \\
 $^3$Beijing National Research Center for Information Science and Technology \\
 $^4$Shanghai Artificial Intelligence Laboratory, Shanghai, China \\
}

\begin{document}
\maketitle
\begin{abstract}
Weakly supervised vision-and-language pre-training~(WVLP), which learns cross-modal representations with limited cross-modal supervision, has been shown to effectively reduce the data cost of pre-training while maintaining decent performance on downstream tasks. 
However, current WVLP methods use only local descriptions of images, i.e., object tags, as cross-modal anchors to construct weakly-aligned image-text pairs for pre-training. This affects the data quality and thus the effectiveness of pre-training. 
In this paper, we propose to directly take a small number of aligned image-text pairs as anchors, and represent each unaligned image and text by its similarities to these anchors, i.e., relative representations. We build a WVLP framework based on the relative representations, namely \modelname\footnote{\modelnamefull}, which collects high-quality weakly-aligned image-text pairs from large-scale image-only and text-only data for pre-training through relative representation-based retrieval and generation. Experiments on four downstream tasks show that \modelname achieves new state-of-the-art results under the weakly supervised setting\footnote{Code is available at \url{https://github.com/THUNLP-MT/RELIT}.}.
\end{abstract}

\section{Introduction}


Vision-and-language pre-training~(VLP)~\cite{chen2020uniter,zhang2021vinvl,kim2021vilt,clip2021radford,wang2022ofa} has received increasing attention in recent years for its great success on various vision-and-language tasks, such as visual question answering~\cite{vqa2015antol}, cross-modal retrieval~\cite{plummer2015flickr30k}, and image captioning~\cite{lin2014coco}. Different from other foundation models~\cite{bommasani2021foundation} such as BERT~\cite{devlin2018bert} and MAE~\cite{MAE2022He} that only require single-modality data, VLP models rely on large-scale aligned image-text datasets~\cite{lin2014coco,sharma2018cc,ordonez2011sbu,krishna2017vg} to bridge the gap between the two modalities, which requires either extensive manual annotations or heavy data cleaning processes~\cite{lin2014coco,sharma2018cc}. 
The natural difficulty of obtaining paired data hinders the scale of cross-modal datasets, while the success of unimodal pre-trained models implies the potential to exploit the unlabeled data for pre-training.
Therefore, besides collecting more paired data, it is a worthwhile direction to explore how to utilize low-cost unimodal data with limited cross-modal supervision, i.e., weakly supervised vision-and-language pre-training~(WVLP).

The core challenge of WVLP is to establish the connection between the two modalities without using a large number of aligned image-text pairs. Existing works on WVLP~\cite{li2021unsupervised,zhou2022unsupervised,wang2022vlmixer,chen2022e2e} usually address this by taking object tags as anchors as they are in the form of text and cover the information of the image at the same time. They use tags to collect weakly-aligned image-text pairs from unaligned unimodal data for pre-training and achieve competitive results compared to standard VLP models, demonstrating that tags can effectively bridge the gap between the two modalities.

Despite its success, using object tags as anchors suffers from two limitations. First, tags are merely local descriptions instead of a complete representation of the whole image and text. Second, the vocabulary of tags only includes common concepts, making it difficult to represent images with complex semantics~\cite{zhou2022unsupervised}. These limitations could deteriorate the quality of the weakly-aligned data (and possibly pre-trained models) based on the object tags. 
Therefore, to further improve the performance of WVLP, we need to reconsider the choice of the cross-modal anchors and find a better approach to measure the alignment between an image and a text.

Recently \textit{relative representation} has been proven to be effective in representation learning~\cite{moschella2022relative} and zero-shot image classification~\cite{norelli2022asif}. The main idea is to represent a data point as its similarities to a set of selected data points (anchors). We argue that relative representations can be a good choice for WVLP because (1) they are built on the semantic similarities of well-trained neural network representations rather than on superficial human-designed features like tags and (2) they are modality-invariant by design because they reflect the intrinsic relationships between data points, which naturally enables communication between different modalities.


In this paper, we propose \modelname, a novel relative representation-based WVLP framework.
Instead of object tags, we directly use a minuscule amount (compared to pre-training data) of available image-text pairs as anchors, and create a common relative representation space with respect to the anchors for unaligned images and text. This allows us to estimate the semantic similarity of any image-text pair by calculating their distance in the relative representation space. In addition, we design two relative representation-based data collection methods that can retrieve or generate weakly-aligned image-text pairs from unaligned unimodal corpora. Experimental results prove the effectiveness of relative representations in bridging the gap between image and text modalities. Moreover, our work reveals a promising research direction to establish cross-modal alignments by finding and aligning invariant data structures in different modalities, which may inspire future works on multimodal pre-training.

Our main contributions are as follows:

\begin{itemize}
    \item We introduce the idea of relative representations in WVLP and demonstrate its superiority over object tags in effectively bridging the gap between different modalities.
    \item We propose a relative representation-based WVLP framework that can both retrieve and generate weakly-aligned image-text pairs for learning cross-modal representations.
    \item Extensive experiments on four diverse vision-and-language tasks show that our proposed framework outperforms strong WVLP baselines and further closes the performance gap between WVLP and standard VLP.
\end{itemize}

\section{Related Work}

\paragraph{Relative Representations.} The concept of relative representations is initially proposed by \citet{moschella2022relative}.
They show that the relative representations obtained from the representation spaces of different models are similar, which enables comparison and alignment between latent embeddings of different learning models. \citet{norelli2022asif} explore relative representations in a multimodal scenario to align images and text for zero-shot image classification tasks. Specifically, they use 1.6M image-text pairs to build the relative representation space, which is comparable to the size of the data used in pre-training. To the best of our knowledge, our work is the first that exploits relative representations for weakly supervised cross-modal pre-training.

\paragraph{Weakly Supervised Vision-and-Language Pre-training.}
\citet{li2021unsupervised} first explore WVLP with unaligned image and text corpora and use object tags directly as pseudo captions for images to bridge the vision and language modalities. \citet{zhou2022unsupervised} use tags to retrieve weakly-aligned captions for each image and then apply multi-granular alignment tasks on this retrieved dataset. \citet{wang2022vlmixer} propose the cross-modal CutMix to replace some grounded words with regions that have the same tags, and construct a multimodal view of the text-only sentences for pre-training. \citet{chen2022e2e} introduce an end-to-end framework with a referring expression matching task. Different from all of these WVLP works that utilize tags as anchors to provide object-level cross-modal alignment signals, our work uses relative representations to capture the overall semantic similarity between each image and text and demonstrates its effectiveness in WVLP.

\paragraph{Data Augmentation.} 
Data augmentation has been extensively employed in various computer vision~\cite{zhang2018mixup,cubuk2018autoaugment} and natural language processing tasks~\cite{sennrich2015backtranslation,guo2020seqmixup}. In the area of VLP, \citet{li2022blip} augment the noisy web-crawled aligned data by filtering low quality image-text pairs and generating synthetic captions with an image captioner fine-tuned on clean image-text pairs. In this work, we adopt a similar filter-and-generate process in the construction of weakly-aligned data for WVLP, but our relative representation-based pseudo caption generator is fine-tuned on the text-only dataset.


\begin{figure*}[htbp]
    \centering
    \begin{subfigure}[b]{0.245\textwidth}
        \centering    
        \includegraphics[width=\textwidth]{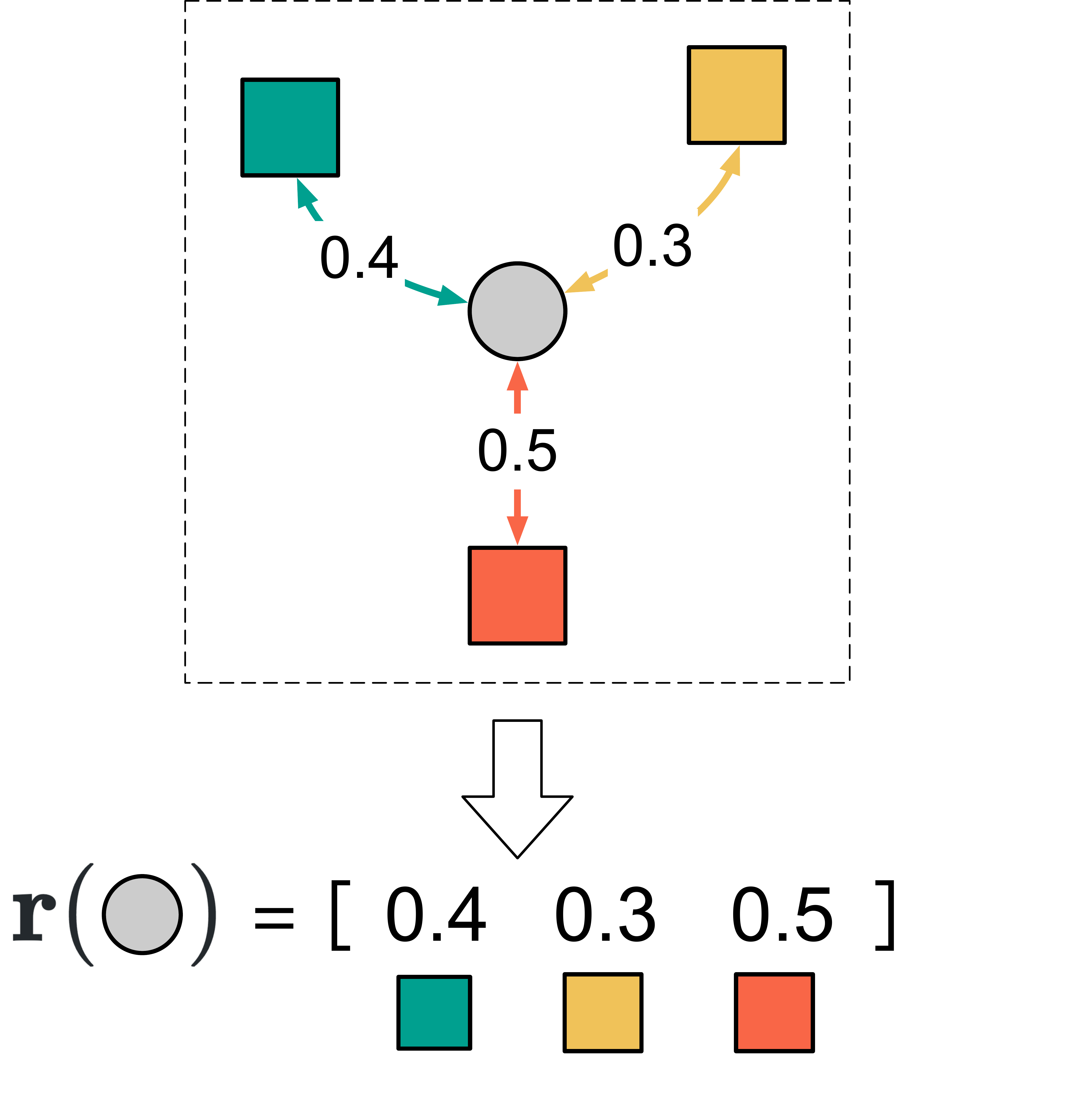}
        \caption{}
        \label{fig:relrep}
    \end{subfigure}%
    \begin{subfigure}[b]{0.745\textwidth}
        \flushright
        \includegraphics[width=0.9\textwidth]{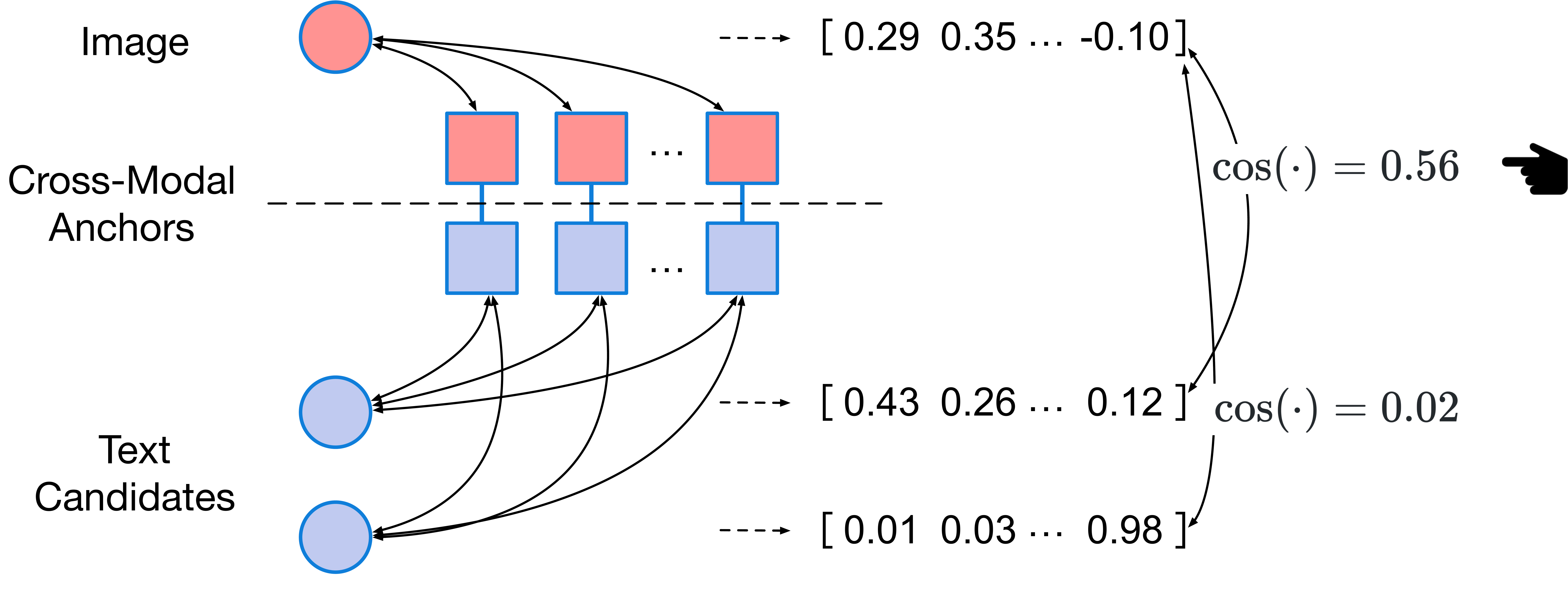}
        \caption{}
        \label{fig:retrieve}
    \end{subfigure}%

    \caption{(a) Illustration of relative representations~(Section~\ref{sec:relrep}), where three anchors~(denoted by squares) are selected and the relative representation of the data point~(denoted by circles) is a 3D vector with each dimension representing its similarity to the corresponding anchor. (b) Image-text retrieval based on the relative representations with cross-modal anchors~(Section~\ref{sec:retrieve}). Data of the same modality are represented by the same color.}
\end{figure*}

\section{Method}

\subsection{Relative Representations}
\label{sec:relrep}

Figure~\ref{fig:relrep} provides an illustration of relative representations. The basic idea is to represent a data point as its similarities to other data points~(anchors). In this work, we consider the relative representations with cross-modal anchors, which has been shown its potential in zero-shot image classification~\cite{norelli2022asif}.

Formally, given a set of $M$ cross-modal anchors $\mathcal{A} = \{a_1, a_2, \ldots, a_{M}\}$ where $a_{i} = (\tilde{x}_{i}, \tilde{y}_{i})$ is an image-text pair, $\tilde{x}_{i}$ is the image and $\tilde{y}_{i}$ is the text. For an image $x$ , a pre-trained image encoder $E_{I}$ is used to calculate the similarity between $x$ and each anchor $a_{i}$ as:
\begin{equation}
    \mathrm{sim}(x, a_{i}) = \cos(E_{I}(x),E_{I}(\tilde{x}_{i}))
\end{equation}
where $\cos(\cdot,\cdot)$ is the cosine similarity, 
and the relative representation of $x$ is defined as:
\begin{equation}
    \mathbf{r_{\mathcal{A}}}(x) = \left( \mathrm{sim}(x, a_{1}), \ldots, \mathrm{sim}(x, a_{M}) \right)
\end{equation}
Similarly, the relative representation of a text $y$ is defined as $\mathbf{r_{\mathcal{A}}}(y)$ with a pre-trained text encoder $E_{T}$ to compute $\mathrm{sim}(y, a_{i}) = \cos(E_{T}(y),E_{T}(\tilde{y}_{i}))$.

Since the relationship between data points is objective, the relative representations obtained by different models should be similar, despite their independent representation spaces~\cite{moschella2022relative}. In other word, an image and its corresponding text should share similar relative representations. This allows us to leverage it to construct weakly-aligned image-text pairs from large-scale unpaired image and text datasets.

\subsection{Weakly-Aligned Image-Text Pairs Retrieval}
\label{sec:retrieve}

While there are no large-scale aligned image-text pairs available, having a joint input of image and text, even if they are not aligned, is still necessary for WVLP~\cite{zhou2022unsupervised,wang2022vlmixer}. To achieve this, inspired by previous work~\cite{zhou2022unsupervised}, we construct a weakly-aligned image-text corpus from the unpaired unimodal corpora by retrieving semantically related sentences for each image based on the relative representations.

Figure~\ref{fig:retrieve} illustrates the process of our weakly-aligned image-text pairs retrieval method. First we collect a very small amount of image-text pairs as cross-model anchors~(denoted by pairs of connected squares in the figure).
Note that the number of anchors is negligible compared to the image-text pairs used in standard VLP, which keeps our method in a weakly supervised setting. 
Then, for all images and text we compute their relative representations with respect to the anchors, which only involves similarity computation within each modality using unimodal pre-trained encoders. We take the cosine distance between the relative representations of each image and text as their semantic relevance score and retrieve the best matching text with the highest score for each image to construct a weakly-aligned image-text pair.

Specifically, we randomly sample $M$ image-text pairs as anchors $\mathcal{A}$ from an aligned image-text dataset~(e.g., Conceptual Captions~\cite{sharma2018cc})~$\mathcal{D}_\text{align}$~($M \ll |\mathcal{D}_\text{align}|$). Given unaligned image dataset $\mathcal{D}_{I}$ and text dataset $\mathcal{D}_{T}$, we construct a retrieved weakly-aligned image-text pair dataset $\mathcal{D}_\text{wa} = \{(x_1, \hat{y}_1), \ldots, (x_N, \hat{y}_N) \}$ where $N = |\mathcal{D}_I|$ and $\hat{y}_i$ is the retrieved caption from $\mathcal{D}_{T}$ for image $x_i$ defined as:
\begin{equation}
    \hat{y}_i = \argmax_{y \in \mathcal{D}_{T}} \cos(\mathbf{r}_{\mathcal{A}}(x_i), \mathbf{r}_{\mathcal{A}}(y))
\end{equation}
We use the off-the-shelf ViT~\cite{dosovitskiy2020vit} and Sentence-BERT~\cite{reimers2019sentencebert} to encode images and text, respectively. 

Our retrieval method with relative representations can effectively improve the quality of retrieved weakly-aligned dataset compared to tag-based retrieval. This is because relative representations tend to capture the overall semantics while tags describe only local information of the image. As a result, our method can better measure the semantic similarities between images and text, especially in cases where tag-based retrieval fails to distinguish between images and text that have different semantics but share the same objects.


\subsection{Pseudo Caption Generation}
\label{sec:generate}

Although relative representation-based retrieval can construct reasonable weakly-aligned image-text pairs for WVLP, there are still cases where non-relevant text are retrieved. This could happen especially when the unaligned unimodal corpora are collected individually and for some images there are no proper captions in the corpora. 

\begin{figure}[t]
\centering 
\includegraphics[width=1.0\textwidth]{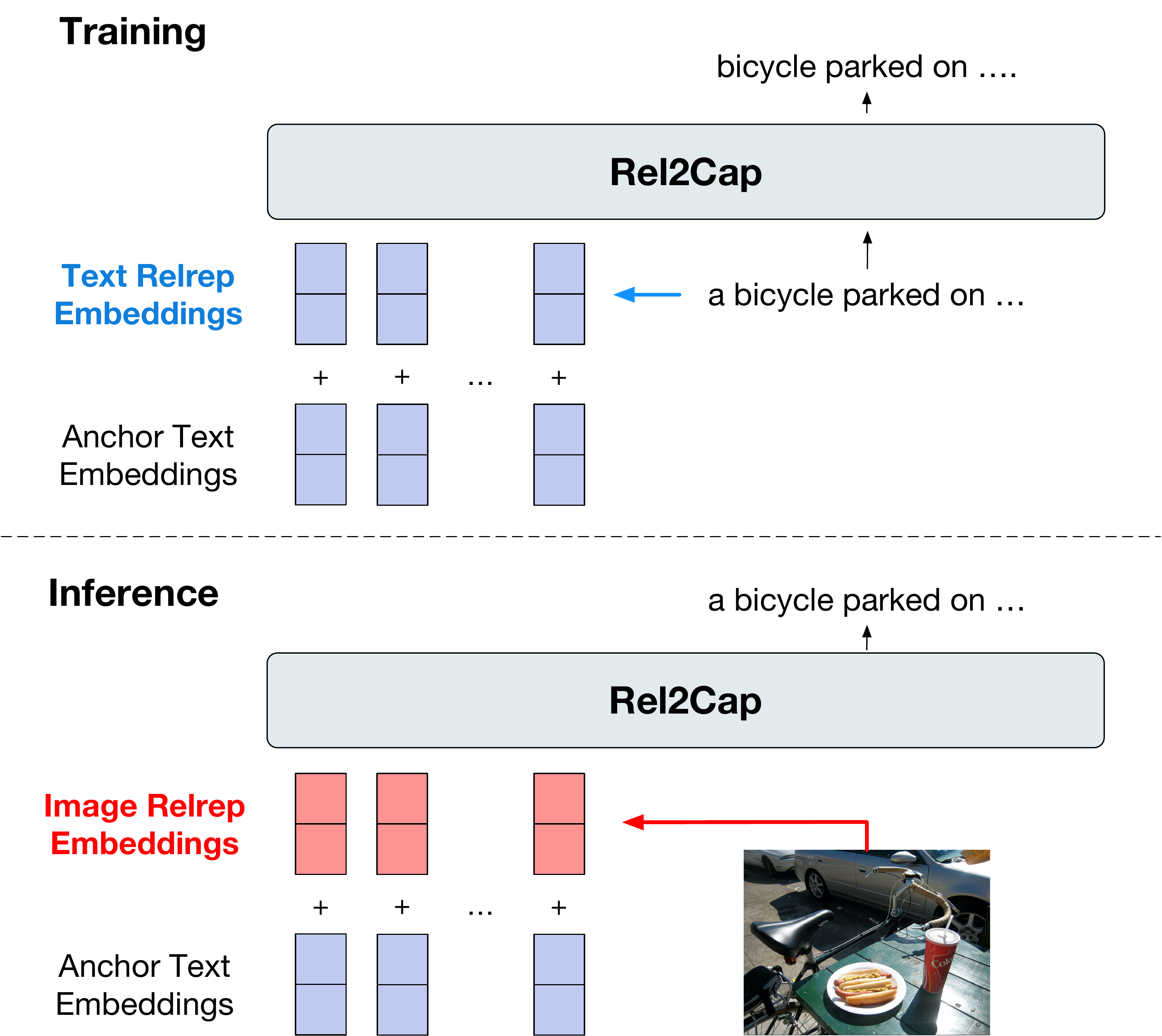}
\caption{An illustration of the training and inference of the pseudo caption generator. In the training process, the model learns to generate text from its relative representation on the text-only dataset. During inference, the model is directly employed to predict the pseudo caption for an image from its relative representation.} 
\label{fig:gen}
\end{figure}

To alleviate this problem, we propose to directly generate pseudo captions for these images. 
As shown in Figure~\ref{fig:gen}, we first adapt a well-trained text generator to perform conditional text generation given relative representations. Then, since images and text share a common relative representation space, we can directly use this generator to predict the pseudo caption for an image based on its relative representation.

Specifically, given the text-only dataset $\mathcal{D}_T$, for each text $y \in \mathcal{D}_T$, we derive a prefix $\mathbf{P} \in {\mathbb{R}^{M\times{d}}}$ from its  relative representations $\mathbf{r}_{\mathcal{A}}(y)$ as:
\begin{equation}
    \mathbf{P} = [\mathbf{r}_{\mathcal{A}}(y)]^{T}\mathbf{W}_{r} + [E_{T}(\tilde{y}_{1}),\ldots,E_{T}(\tilde{y}_{M})]\mathbf{W}_{e}
\label{eq:rel2cap}
\end{equation}
where $E_{T}(\tilde{y}_{i}) \in \mathbb{R}^{d_T}$ is the encoder output of the text in the $i$-th anchor, $\mathbf{W}_{r} \in \mathbb{R}^{1\times{d}}$ and $\mathbf{W}_{e}  \in \mathbb{R}^{d_T\times{d}}$ are two learnable projection matrices.
We fine-tune a pre-trained GPT-2 model~\cite{radford2019gpt2} to learn to predict $y$ given $\mathbf{P}$, and name the fine-tuned model as \textit{Rel2Cap}. To further save computational cost, we only consider the entries in $\mathbf{P}$ that correspond to the top $K$ anchors with the highest similarities as the model input.

After training, the model can be used to predict the pseudo caption for an image $x$ with low quality retrieved captions by constructing an input prefix $\mathbf{P}^{'}$ based on the relative representations of the image, i.e., $\mathbf{r}_{\mathcal{A}}(x)$.
The definition of $\mathbf{P}^{'}$ is similar to Equation~\ref{eq:rel2cap}, except that $\mathbf{r}_{\mathcal{A}}(y)$ is replaced by $\mathbf{r}_{\mathcal{A}}(x)$. 
We define a quality score $\mathrm{s}(x, \hat{y}) = \cos(\mathbf{r}_\mathcal{A}(x), \mathbf{r}_\mathcal{A}(\hat{y}))$ for each weakly-aligned image-text pair $(x, \hat{y})$ collected both by retrieval and generation, and replace the retrieved pair with the generated one if the latter has a higher quality score. 

So far, we have discussed how we collect a weakly-aligned image-text dataset $\mathcal{D}_{\text{wa}}$ from the unpaired unimodal corpora by relative representation-based retrieval and generation. Next, we describe how we use these data for WVLP. 


\subsection{Pre-training}

\paragraph{Model Overview.}

We use the same model architecture as \citet{chen2022e2e} that consists of a vision and a multimodal encoder. For each weakly-aligned image-text pair, the image is encoded with the vision encoder and the outputs are fed to the multimodal encoder along with the text embeddings to obtain a multimodal representation. Such an end-to-end framework has been proven to be more effective compared to others that use region features from external object detectors both in standard VLP and WVLP. 
We apply three pre-training objectives to learn multimodal representations from the collected weakly-aligned image-text pairs: masked tag prediction~(MTP), masked language modeling~(MLM) and image text matching~(ITM). 

\paragraph{Masked Tag Prediction.} This objective aims to learn object-level cross-modal alignment from the image-only data and their detected object tags. Following previous works~\cite{li2021unsupervised, chen2022e2e}, we randomly mask out the tags with a probability of $15\%$, and then predict the masked tags conditioned on the image and other unmasked tags. Formally, given the image $x \in \mathcal{D}_{I}$ and its detected object tags $t$, the MTP objective is defined as:
\begin{equation}
    \mathcal{L}_{\text{MTP}} = -\mathbb{E}_{x\in\mathcal{D}_{I}}\log P(t_m|t_{\backslash m},x)
\end{equation}
where $t_m$ and $t_{\backslash m}$ represents masked and unmasked object tags, respectively.

\paragraph{Masked Language Modeling.} To better fuse between the two modalities, the masked language modeling objective is adopted to learn from the joint image-text inputs from the weakly-aligned corpora. Since the weakly-aligned pairs may contain noise in the retrieved or generated text, we only mask out and predict the noun phrases in the text inspired by \cite{zhou2022unsupervised}. The MLM loss is formulated as:
\begin{equation}
    \mathcal{L}_{\text{MLM}} = -\mathbb{E}_{(x,\hat{y})\in\mathcal{D}_{\text{wa}}}\log P(\hat{y}_m|\hat{y}_{\backslash m},x)
\end{equation}
where $\hat{y}_m$ and $\hat{y}_{\backslash m}$ are masked and unmasked text.

\paragraph{Image Text Matching.} ITM is a commonly used objective for learning instance-level cross-modal alignment in VLP, which aims to distinguish whether an image-text pair is matched semantically. We random replace the text in half of the image-text pairs with another text to form training input, and define the label of each pair as $l\in\{0,1\}$ where $1$ indicates the pair is a match. The ITM objective is to minimize the binary cross-entropy loss:
\begin{equation}
    \mathcal{L}_{\text{ITM}} = -\mathbb{E}_{(x,\hat{y})\in\mathcal{D}_{\text{wa}}^{'}}\log P(l|x,\hat{y})
\end{equation}
where $\mathcal{D}_{\text{wa}}^{'}$ is the dataset after random replacement.

\paragraph{Relative Representation-Guided Training.}

To further reduce the impact of the noisy image-text pairs in the weakly-aligned dataset, we apply the quality score $\mathrm{s}(x, \hat{y})$ of each pair described in Section~\ref{sec:generate} to $\mathcal{L}_{\text{MLM}}$ and $\mathcal{L}_{\text{ITM}}$ to guide the training to learn more from high-quality data:
\begin{align}
    \mathcal{L}_{\text{MLM}} &= -\mathbb{E}_{(x,\hat{y})\in\mathcal{D}_{\text{wa}}}\mathrm{s}(x, \hat{y})\log P(\hat{y}_m|\hat{y}_{\backslash m},x) \\
    \mathcal{L}_{\text{ITM}} &= -\mathbb{E}_{(x,\hat{y})\in\mathcal{D}_{\text{wa}}^{'}}\mathrm{s}(x, \hat{y})\log P(l|x,\hat{y})
\end{align}

\begin{table*}[!ht]\centering
\small
\begin{tabular}{l|c|c|c|ccc}\toprule
\multirow{2}{*}{\bf Model }  & {\bf VQAv2} & {\bf \nlvr} & {\bf VE} & \multicolumn{3}{c}{\bf Flickr30k} \\
& {\bf Test-Dev} & {\bf Test-P} & {\bf Test} & {\bf R@1} & {\bf R@5} & {\bf R@10} \\
\midrule
\multicolumn{7}{c}{\it Supervised (w/ Large-Scale Paired Image-Text Data)}\\
\midrule
VisualBERT~\citep{li2019visualbert} & 70.9 &73.9 &- & 61.2 & 86.3 & 91.9 \\
UNITER~\citep{chen2020uniter} & 72.7 & 77.9 & 78.3 & 72.5 & 92.4 & 96.1 \\
VinVL~\citep{zhang2021vinvl} & 76.0 & \textbf{83.1} & - & - & - & - \\
ViLT~\citep{kim2021vilt} & 71.3 & 76.1 & - & 66.4 & 88.7 & 93.8 \\
ALBEF~\citep{li2021align} & 74.5 & 80.5 & 80.3 & \textbf{82.8} & \textbf{96.7} & \textbf{98.4} \\
\textsc{Meter}\xspace-CLIP-ViT$_{\text{BASE}}~$\citep{dou2022empirical} & \textbf{77.7} & 83.0 & \textbf{81.2} & 82.2 & 96.3 & 98.3 \\
\midrule
\multicolumn{7}{c}{\it Weakly Supervised (w/o Large-Scale Paired Image-Text Data)}\\
\midrule
$\text{\uvbert}$~\citep{li2021unsupervised} & 70.7 &71.0 & - & 55.4 & 82.9 & 89.8 \\
$\text{\uvbert}_{\text{VinVL}}$~\citep{zhou2022unsupervised} & 71.8 & 53.2 &76.8 & - & - & -\\
$\text{\muvla}$~\citep{zhou2022unsupervised} & 72.1 & 73.4 & 77.3 & - & - & - \\
$\text{\vlmixer}$~\citep{wang2022vlmixer} & 72.7 & 73.9 & - &  - & - & - \\
$\text{\eteuvlp}$~\citep{chen2022e2e} & 73.3 & 74.6 & 78.2 & 66.4 & 89.7 & 94.1  \\\midrule
$\text{\modelname~(Ours)} $ & \textbf{73.5} & \textbf{76.4} & \textbf{78.6} & \textbf{70.2} & \textbf{91.5} & \textbf{95.6} \\
\bottomrule
\end{tabular}
\caption{Evaluation results on four V+L downstream tasks. All weakly-supervised models are pre-trained on non-parallel images and text from CC.}
\label{tab:main}
\end{table*}

\begin{table*}[!ht]\centering
\small
\begin{tabular}{l|c|c|c|ccc}\toprule
\multirow{2}{*}{\bf Method }  & {\bf VQAv2} & {\bf \nlvr} & {\bf VE} & \multicolumn{3}{c}{\bf Flickr30k} \\
 & {\bf Test-Dev} & {\bf Test-P} & {\bf Test} & {\bf R@1} & {\bf R@5} & {\bf R@10} \\\cmidrule{1-7}
\uvbert & 70.5 & 71.2 & - & 54.4 & 82.2 & 89.2 \\
\muvla & 71.2 & 67.1 & 77.1 & - & - & - \\
\eteuvlp & 73.5 & 73.7 & 77.9 & 65.6 & 90.3 & 94.7 \\ \midrule
\modelname~(Ours) & \textbf{73.6} & \textbf{74.8} & \textbf{78.2} & \textbf{67.7} & \textbf{90.4} & \textbf{95.0} \\
\bottomrule
\end{tabular}
\caption{Experimental results on downstream tasks of pre-training with images from CC and text from BookCorpus.}
\label{tab:bc}
\end{table*}

\section{Experiments}
\subsection{Datasets}

We follow previous WVLP works~\cite{li2021unsupervised,zhou2022unsupervised,wang2022vlmixer,chen2022e2e} and conduct experiments in two different settings, each containing an image-only dataset and a text-only dataset. The first setting treats images and text from Conceptual Captions~(CC)~\cite{sharma2018cc} as individually collected unimodal dataset without the alignment information. The second setting uses images from CC and text from BookCorpus~\cite{zhu2015bc}, which is a more realistic scenario where images and text are gathered separately from different sources.




\subsection{Implementation Details}
\paragraph{Relative Representations.} We randomly select $8,192$ aligned image-text pairs from CC as anchors, yielding relative representations as vectors of $8,192$ dimensions. To save computational cost, inspired by \citet{norelli2022asif}, we only keep the highest $50$ dimensions and set the others to $0$.

\paragraph{Weakly-Aligned Data Construction.} We implement the retrieval system with the faiss~\cite{johnson2019faiss} library. For each image we only retrieve the text with the best match score. For RelCap, we fine-tune GPT-2 with a learning rate of 5e-5 and a batch size of $1,024$ for $5$ epochs on the text-only dataset. We generate $5$ pseudo-captions for each image using nucleus sampling with $p=0.9$ which proved effective in synthetic caption generation~\cite{li2022blip}, and rank the results with the quality scores. We also include the weakly-aligned dataset based on tag-based retrieval in the pre-training, as described in \citet{zhou2022unsupervised}.

\paragraph{Pre-training.} We use the same architecture as \citet{chen2022e2e} which includes a 12-layer Swin-Transformer~(Swin B-384/32)~\cite{liu2021swin} as the vision encoder and a 12-layer Transformer initialized from BERT-base~\cite{devlin2018bert} as the multimodal encoder. For object tags, we utilize the off-the-shelf object detector provided by VinVL~\cite{zhang2021vinvl}. We pre-train the model with a total training step of $150$k and a batch size of $512$. We use an AdamW optimizer~\cite{kingma2014adam} with an initial learning rate of 3e-5, and the warm-up ratio is set to $10\%$. The pre-training takes $3$ days on $4$ NVIDIA A100 GPUs. 

\paragraph{Downstream Tasks.} We follow previous works and test our pre-trained model on four downstream V+L tasks, including Visual Question Answering~(VQAv2)~\cite{balanced_vqa_v2}, Natural Language for Visual Reasoning~(\nlvr)~\cite{suhr2018corpus}, Visual Entailment~(VE)~\cite{xie2019visual} and image retrieval~(Flickr30k)~\cite{plummer2015flickr30k}. Details of the task settings and the fine-tuning strategies are in Appendix~\ref{sec:downstream}.

\subsection{Main Results}

We first compare our proposed \modelname with previous methods pre-trained with unaligned images and text from CC. Note that these baselines only utilize object tags. Table~\ref{tab:main} shows the experimental results on the downstream tasks. Our method outperforms previous WVLP methods on all downstream tasks. Specifically, \modelname outperforms previous best results by $1.8\%$ on \nlvr and by $3.8\%$ on the image retrieval task (Flickr30k), both of which benefit from the instance-level cross-modal alignment capability of the pre-trained model~\cite{chen2020uniter,zhou2022unsupervised}. This suggests that our relative representation-based method improves the alignment quality of weakly-aligned image-text pairs compared to previous tag-based approaches, resulting in improved cross-modal alignment capability of the pre-trained model. 

When pre-trained with images from CC and text from BookCorpus, as shown in Table~\ref{tab:bc}, our proposed \modelname also achieves the best results on all downstream tasks. This demonstrates that the proposed relative representation-based methods can effectively mine useful cross-modal alignment information for multimodal pre-training from image-only and text-only data, even if they are collected separately from different sources.

\subsection{Ablation Study}

\begin{table*}[!ht]\centering
\small
\begin{tabular}{l|c|c|c|ccc}\toprule
\multirow{2}{*}{\bf Pre-training Data } & {\bf VQAv2} & {\bf \nlvr} & {\bf VE} & \multicolumn{3}{c}{\bf Flickr30k} \\
 & {\bf Test-Dev} & {\bf Test-P} & {\bf Test} & {\bf R@1} & {\bf R@5} & {\bf R@10} \\\cmidrule{1-7}
\tagretrv & 73.2 & 74.5 & 77.8 & 66.3 & 89.3 & 94.2 \\
\relrepretrv & 73.4 & 74.9 & 78.3 & 67.5 & 90.5 & 94.9 \\
\tagretrv + \relrepretrv & \textbf{73.5} & 75.3 & 78.4 & 67.3 & 90.4 & 94.6 \\
\tagretrv + \relrepretrv + Rel2Cap & \textbf{73.5} & \textbf{76.4} & \textbf{78.6} & \textbf{70.2} & \textbf{91.5} & \textbf{95.6} \\
\bottomrule
\end{tabular}
\caption{Comparison of pre-training with different kinds of pseudo-aligned data.}
\label{tab:ablation}
\end{table*}

\begin{table}[!ht]\centering
\small
\begin{tabular}{l|c|c|c|c}\toprule
\multirow{2}{*}{\bf Method } & {\bf VQAv2} & {\bf \nlvr} & {\bf VE} & {\bf Flickr30k} \\
 & {\bf Test-Dev} & {\bf Test-P} & {\bf Test} & {\bf R@1} \\\cmidrule{1-5}
\modelname & 73.5 & 76.4 & 78.6 & 70.2 \\
- Guided & 73.2 & 76.1 & 78.5 & 70.0  \\
\bottomrule
\end{tabular}
\caption{Ablation study on relative representation-guided training.}
\label{tab:guide}
\end{table}

We conduct an ablation study to verify the effectiveness of the proposed relative representation-based retrieval and generation. Table~\ref{tab:ablation} shows the results. All models are pre-trained on weakly-aligned data derived from unaligned CC images and text.
As we can see from the table, compared to tag-based retrieved data~(\tagretrv), pre-training with relative representation-based retrieved data (\relrepretrv) performs better on downstream tasks. Besides, the model achieves the best results when the generated pseudo captions (Rel2Cap) are included during pre-training. We believe this is because the original CC dataset contains noisy captions, such as alt-texts that do not describe the image contents, which is suboptimal for VLP~\cite{li2022blip}. In summary, the experimental results demonstrate that both our retrieval and generation methods contribute to the performance of the pre-training.

We also compare the performance of the pre-trained models on downstream tasks with and without relative representation-guided training. As shown in  Table~\ref{tab:guide}, pre-training with guided training can consistently improve results across all downstream tasks, illustrating that relative representations can be used to detect noise in the weakly-aligned data and guide the model to learn from data with a higher level of alignment.


\subsection{Data Quality}

We evaluate the quality of different kinds of weakly-aligned data from unaligned CC images and text, and the results are listed in Table~\ref{tab:data_quality}. We use CLIPScore~\cite{hessel2021clipscore} to measure the overall alignment of all weakly-aligned image-text pairs. As we can see from the table, the data quality of \relrepretrv is significantly higher than that of \tagretrv, which again illustrates the superiority of relative representations as cross-modal anchors. In addition, Rel2Cap further improves data quality by filtering and replacing low-quality pairs in \relrepretrv. The analysis of the data quality here is consistent with the analysis of pre-training results in Table~\ref{tab:ablation}, and again proves that our relative representation-based methods can produce high quality weakly-aligned data from unaligned unimodal data.

\begin{table}[t]\centering
\small
\begin{tabular}{l|c}\toprule
{\bf Data } & {\bf CLIPScore} \\
\cmidrule{1-2}
\tagretrv & 57.94 \\
\cmidrule{1-2}
\relrepretrv & 63.31 \\
+ Rel2Cap & 65.23 \\
\bottomrule
\end{tabular}
\caption{Data quality of different kinds of weakly-aligned data from unaligned CC images and text.}
\label{tab:data_quality}
\end{table}

\subsection{Effects of Anchor Selection}

\begin{figure}[htbp]
\centering 
\includegraphics[width=1.0\textwidth]{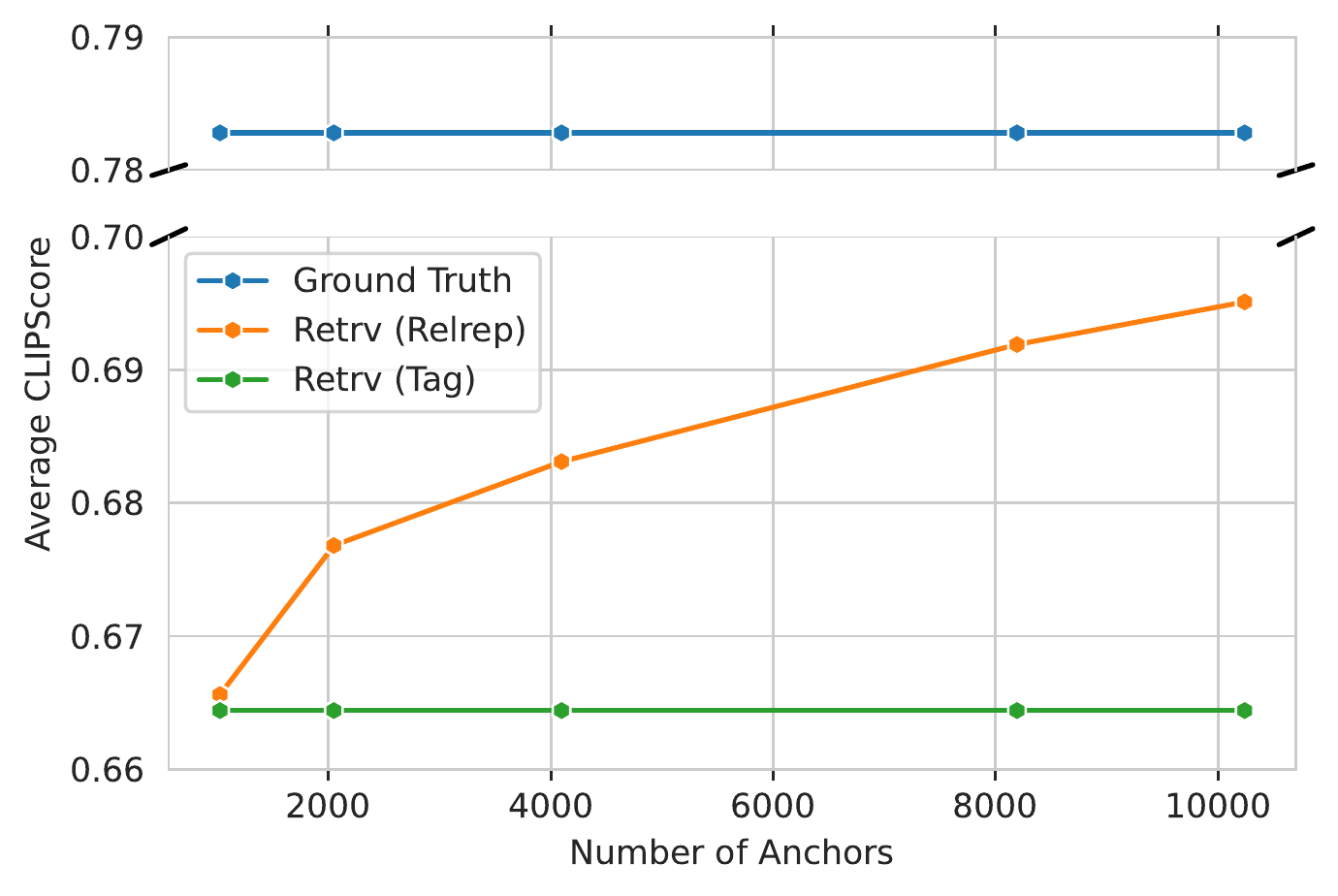}
\caption{Data quality of the retrieved data using different number of anchors. Both the anchors and the images and text used for retrieval are from the COCO dataset.} 
\label{fig:anchor_clipscore}
\end{figure}

\begin{figure}[htbp]
\centering 
\includegraphics[width=1.0\textwidth]{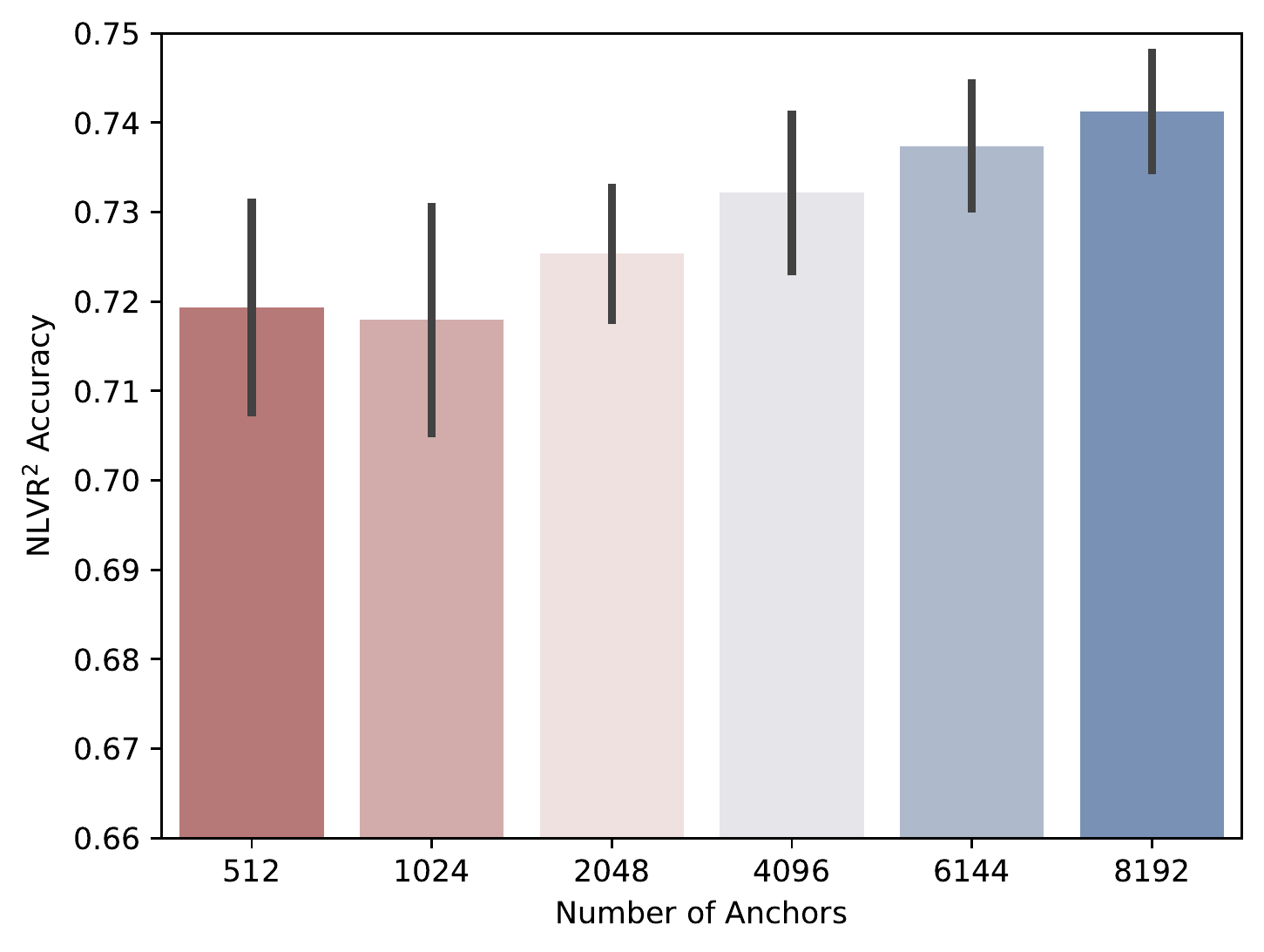}
\caption{Fine-tuned \nlvr results of models pre-trained on data with different number of anchors.} 
\label{fig:anchor_nlvr2}
\end{figure}

The number of anchors has a significant influence on the effect of relative representations~\cite{norelli2022asif}. To verify its influence on the collected weakly-aligned image-text pairs, we test the quality of the data retrieved with different numbers of anchors on the COCO~\cite{lin2014coco} dataset. From Figure~\ref{fig:anchor_clipscore}, we can see that as the number of anchors increases, the quality of the retrieved data also improves. In addition, we evaluate the downstream task accuracy using models pre-trained on data with varying numbers of anchors. Specifically, we generate 3 random sets of anchors for each size, and retrieve the weakly-aligned data with different sets of anchors. We pre-train models on each set of the retrieved data with the same hyperparamters, and fine-tune them on the \nlvr task. The results are shown in Figure~\ref{fig:anchor_nlvr2}. In general, the higher the number of anchors, the better the model performance.
We use $8,192$ anchors in our final experiments as a trade-off between representation capability and computational cost. However, using more anchors will almost certainly give better results due to better quality of the data, which indicates the scalability of our approach. We leave more exploration on this for future work.

\begin{table}
\centering
\small
\begin{tabular}{lcccc}
\toprule
\multirow{2}{*}{\textbf{Method}} & \multicolumn{4}{c}{\textbf{Number of Anchors}} \\
\cmidrule{2-5}
 & \textbf{1024} & \textbf{2048} & \textbf{4096} & \textbf{8192} \\
\midrule
random & 66.50 & 67.68 & 68.31 & 69.19 \\
diverse & 67.66 & 68.34 & 68.60 & 69.37 \\
non-diverse & 40.94 & 39.52 & 39.53 & 39.50 \\
\bottomrule
\end{tabular}
\caption{The quality of data obtained from different anchor selection methods.}
\label{tab:anchor_sampling}
\end{table}

We also conduct experiments to verify the impact of anchor diversity on data quality. Specifically, we considered three sampling methods on the COCO dataset: \textit{random}, \textit{diverse}, and \textit{non-diverse}. The diverse sampling first performs K-means clustering on all the data, and selects one anchor from each cluster. The non-diverse sampling uses a greedy algorithm to select k anchors, at each step choosing the data closest to the average of the anchors already selected.  Table~\ref{tab:anchor_sampling} lists the data quality results obtained with different sampling methods. In general, diverse anchors lead to better quality, while random anchors perform satisfactorily when the number of anchors is large enough. Non-diverse anchors can result in catastrophic data quality.


\subsection{Case Study}


\begin{figure*}[htbp]
\centering 
\includegraphics[width=1.0\textwidth]{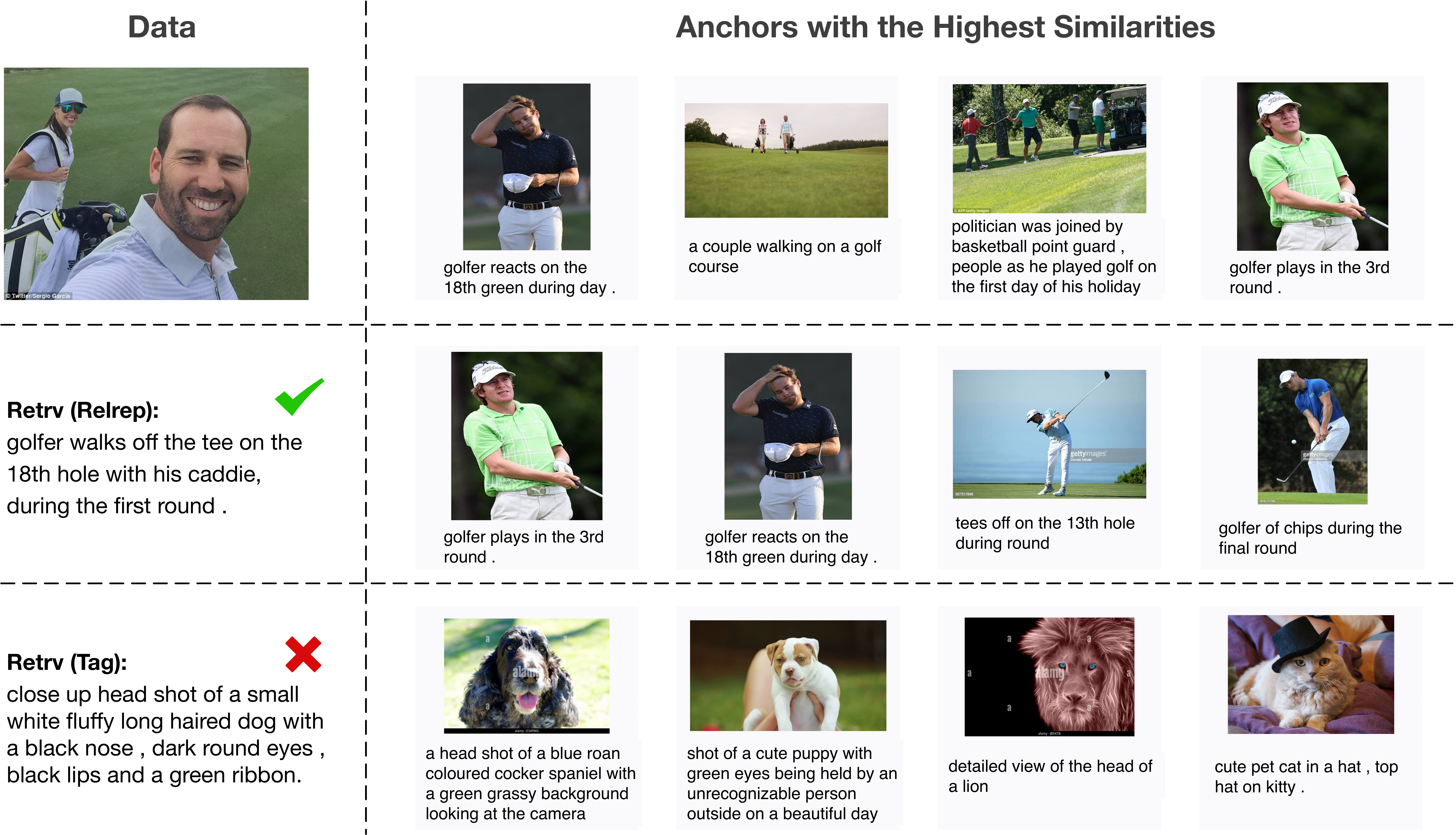}
\caption{Comparison of the relative representations of the image and retrieved captions in Figure~\ref{fig:case_a}. For simplicity, for each image and text on the left, we only display the anchors with the highest similarities on the right. } 
\label{fig:anchor}
\end{figure*}

\begin{figure}[!t]
\centering
\begin{subfigure}[b]{\textwidth}
        \centering
        \includegraphics[width=\textwidth]{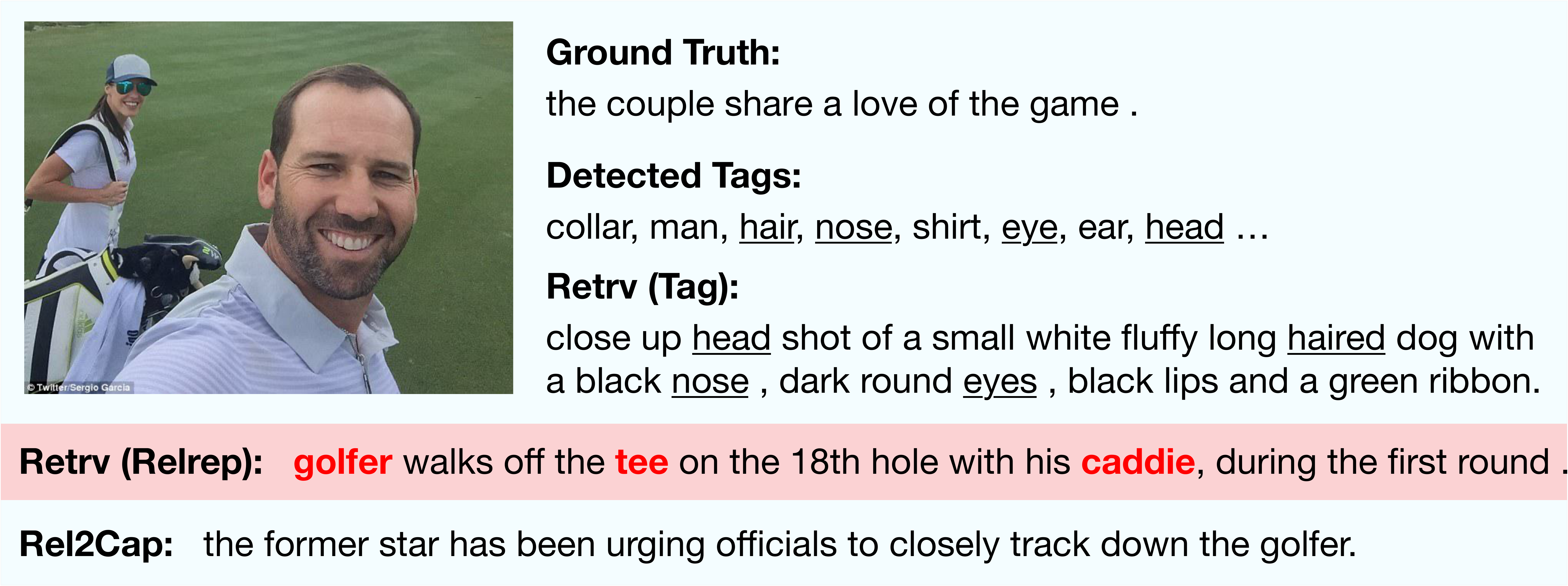} 
        \caption{}
        \label{fig:case_a}
\end{subfigure}
\begin{subfigure}[b]{\textwidth}
        \centering
        \vspace{0.4em}
        \includegraphics[width=\textwidth]{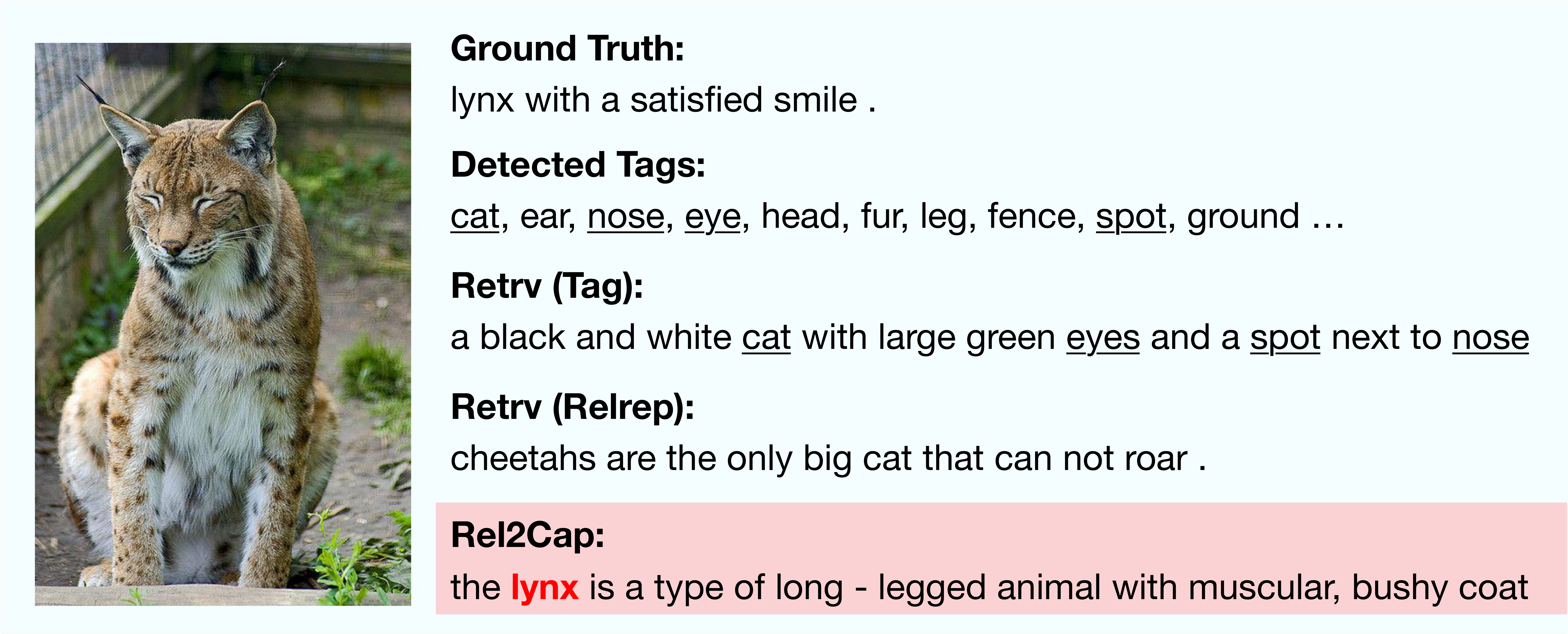} 
        \caption{}
        \label{fig:case_b}
\end{subfigure}
\caption{Examples of different kinds of weakly-aligned data. We highlight in red the caption with the best quality and the words in it that match the key information of the image. Compared to \tagretrv which focuses on tag matching~(underlined), our proposed two methods \tagretrv and Rel2Cap produce captions that are more semantically similar to the image.}
\label{fig:case}
\end{figure}

To explore the reasons for the improvement in data quality, we show two examples of the comparisons between different weakly-aligned image-text pairs in Figure~\ref{fig:case}. In each example, we provide the ground truth caption of the image and the detected object tags, as well as three weakly-aligned captions. From these two examples, we can see that the captions retrieved by tags do have many of the same tags as the images (underlined in the figure), but are not good descriptions of the images. In contrast, our relative representation-based retrieval and generation methods are able to obtain captions that are more relevant to the overall semantics of the images. Specifically, in the example in Figure~\ref{fig:case_a}, our proposed methods successfully identifies key information in the image such as ``golfer'', which is difficult for tag-based retrieval since there are no such tag as ``golfer''. The same thing happens to \tagretrv in Figure~\ref{fig:case_b}, which retrieves a caption related to ``cat'' instead of ``lynx''. In this example, our retrieval method recognizes the animal in the image as ``cheetah'', which is close but not exactly correct, while our generation method correctly generates a caption related to the correct concept ``lynx''. This indicates that our generation method has the ability to generate pseudo captions of better quality when the retrieved ones are not good enough.

In Figure~\ref{fig:anchor} we further visualize the relative representations of the image and two retrieved captions in Figure~\ref{fig:case_a}, which helps understand the effectiveness of relative representations in aligning semantically related image-text pairs. From the figure we can see that the image and our retrieved caption \relrepretrv activate the same group of anchors~(i.e., have high similarities with these anchors), which makes them close in the relative representation space. On the other hand, \tagretrv activates a completely different set of anchors, which leads to a large distance between it and the image in the relative representation space. These observations suggest that (1) relative representations are (almost) modality-invariant and (2) relative representations can be utilized to effectively estimate the cross-modal alignment of data in different modalities. These properties of the relative representations make it naturally suitable for WVLP, which is verified in this paper.

\section{Conclusion}

This paper introduces the idea of relative representations to weakly-supervised vision-and-language pre-training and demonstrates its effectiveness in bridging the gap between the two modalities. We propose a relative representation-based framework that can both retrieve and generate weakly-aligned image-text pairs for pre-training. Experimental results show that our method outperforms all previous tag-based approaches under the weakly-supervised setting. We hope our work will motivate future work in multimodal pre-training. 

\section*{Limitations}

As this work is mainly focused on weakly supervised vision-and-language pre-training, we do not fully explore the factors that may influence the performance of relative representations, such as the use of different unimodal encoders and the source of the anchors. Besides, we only validate the effectiveness of relative representations in a weakly supervised setting, while it remains to be explored whether it is also useful for standard VLP and multimodal learning in other modalities~(e.g., audio and video). We will further exploit the potential of relative representations and validate it in more cross-modal learning scenarios in the future.

\section*{Acknowledgments}

This work is supported by the National Key R\&D Program of China (2022ZD0160502) and the National Natural Science Foundation of China (No. 61925601, 62276152, 62236011). We thank Ziyue Wang, Fuwen Luo, Rui Jiao and Zonghan Yang for their advices in paper writing.

\bibliography{custom}
\bibliographystyle{acl_natbib}

\clearpage
\appendix

\section{Details of Downstream Tasks}
\label{sec:downstream}

\paragraph{Visual Question Answering~(VQA)} The task of VQA is to answer questions correctly according to the given images. We follow previous works~\cite{yu2019deep,chen2020uniter} and formulate VQA as a classification task with $3,192$ classes representing the most frequent answers in the dataset. We fine-tune the pre-trained model for $10$ epochs with a batch size of $256$. We use an AdamW optimizer with a peak learning rate of $5\times 10^{-5}$.

\paragraph{Natural Language for Visual Reasoning (\nlvr)} The objective of \nlvr is to decide if a natural language description is true for a given pair of images. We follow previous work~\cite{chen2020uniter} to form two image-text pairs as inputs, and concatenate the two [CLS] outputs of the model as the final representation for classification. We fine-tune the model for $10$ epochs with a batch size of $128$ and a peak learning rate of $2.5\times 10^{-5}$.

\paragraph{Visual Entailment~(VE)} Given an image and a text hypothesis, the task of VE is to determine whether the image implies the hypothesis. This is formulated as a three-way classification task to predict whether the logical relationship between the image and the text is \textit{entailment}, \textit{neutral} or \textit{contradiction}. For the VE task, we fine-tune the pre-trained model with a batch size of $64$ and a peak learning rate of $1\times 10^{-5}$ for $5$ epochs. 

\paragraph{Image Retrieval~(Flickr30k)} We follow previous works~\cite{li2021unsupervised,chen2022e2e} to conduct the image retrieval task on the Flickr30k~\cite{plummer2015flickr30k} dataset. We sample $15$ negative image-text pairs for each positive pair by replacing its text with randomly sampled ones. The batch size is set to $512$. We fine-tune the model with a peak learning rate of $2.5\times 10^{-5}$ for $10$ epochs.

\begin{figure*}[t]
\centering 
\includegraphics[width=1.0\textwidth]{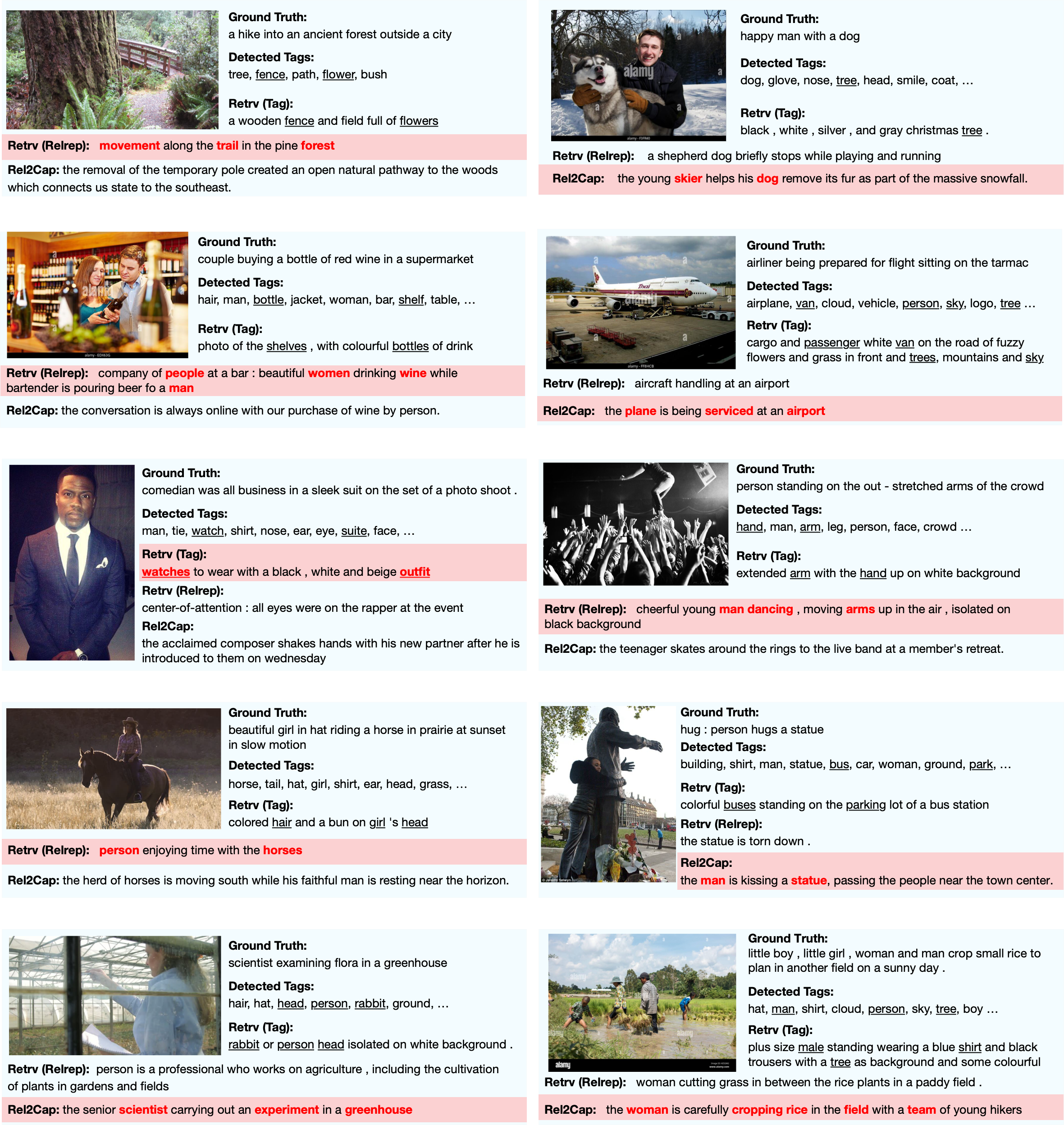}
\caption{Examples of different kinds of weakly-aligned data. We highlight in red the caption with the best quality and the words in it that match the key information of the image.} 
\label{fig:case_appendix}
\end{figure*}

\section{Additional Examples}

In Figure~\ref{fig:case_appendix}, we provide more examples of different kinds of weakly-aligned image-text pairs. From these examples, we can see that our relative representation-based approaches yield higher quality weakly-aligned image-text pairs compared to tag-based retrieval.

\end{document}